\documentclass[letterpaper]{article}
\usepackage[utf8]{inputenc}
\usepackage{natbib,alifeconf}
\usepackage{graphicx}
\PassOptionsToPackage{hyphens}{url}\usepackage{hyperref}

\graphicspath{{figures/}}

\title{Perspective: Purposeful Failure in Artificial Life and Artificial Intelligence}
\author{Lana Sinapayen\\
\mbox{}\\
Sony Computer Science Laboratories, Kyoto Laboratory, 13-1 Hontorocho, Shimogyo Ward, Kyoto 6008086 Japan \\
lana.sinapayen@gmail.com}

\begin{document}
\maketitle

\begin{abstract}
Complex systems fail. I argue that failures can be a blueprint characterizing living organisms and biological intelligence, a control mechanism to increase complexity in evolutionary simulations, and an alternative to classical fitness optimization. Imitating biological successes in Artificial Life and Artificial Intelligence can be misleading; imitating failures offers a path towards understanding and emulating life it in artificial systems. 
\end{abstract}

\section{Failure is Knowledge, Knowledge is Power}

You are handed a mysterious box containing the most complex object in the universe, and must find how the object works. Where do you start?

``The human brain is the most complex object in the universe” is a well worn cliche (\cite{challice}). While the claim might not be true, the human brain is definitely very complex. In neuroscience and psychology, one of the most compelling ways to understand how the brain works is to study how it fails. Brain damage, irrational decisions, sensory illusions: internal or external changes that make the brain fail are how we find how the brain succeeds. Failure is used to understand complex systems beyond neuroscience: reverse-engineering computer software, understanding animal behavior, identifying solid materials... Failure even defines Science itself. an hypothesis is considered scientific if and only if it is ``falsifiable": if it can reproducibly fail (\cite{Popper}).

\begin{figure}[!htb]
\begin{center}
\includegraphics[width=\linewidth]{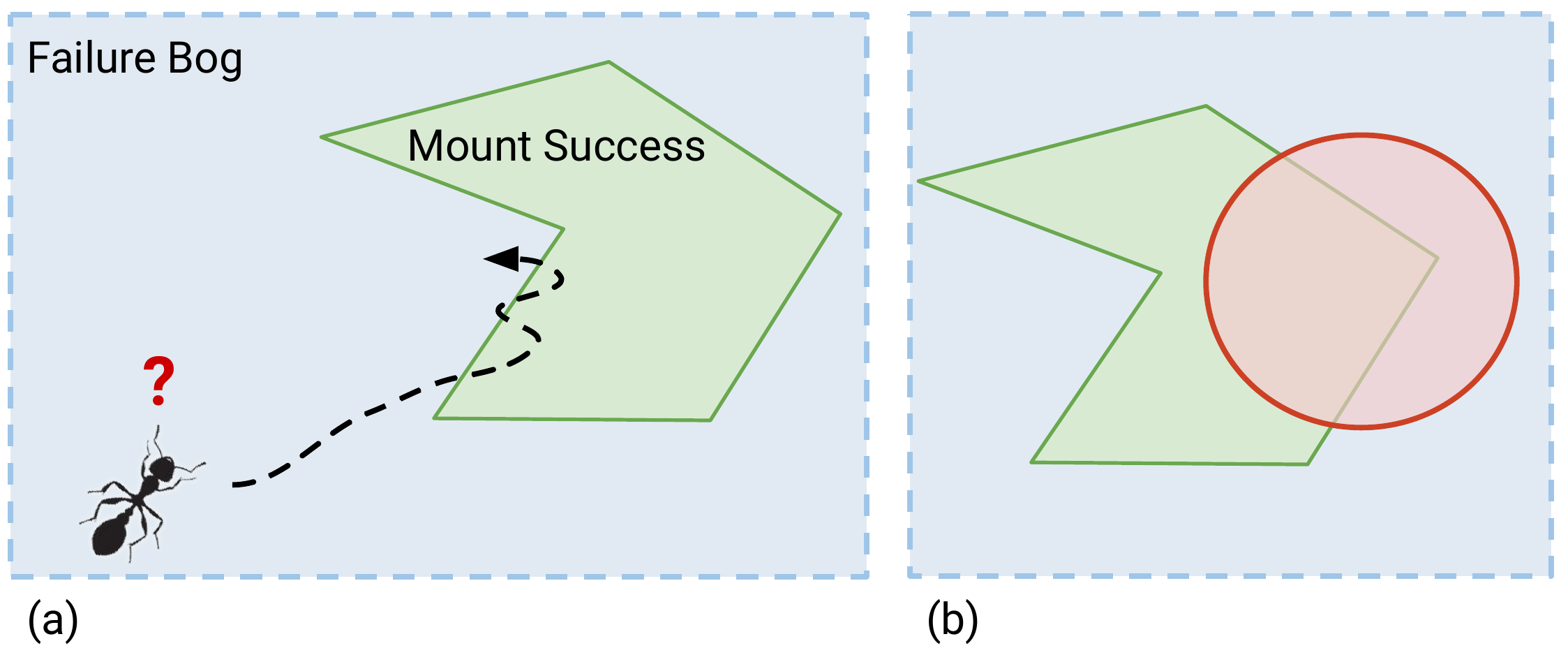}
\caption{\textbf{(a) The fastest way to map this bog-and-mount terrain is to find the boundary and walk along it.} Similarly, a complex system can be characterized by its boundary between failure and success. Soft transitions can be hard to notice, so we are better off finding a few remarkable cliffs off Mount Success, abrupt ``failure points" transitions from success to failure, and from there extrapolating to the whole failure boundary.
\textbf{(b) The green system and the red system share many successes, but only 2 points of their failure boundary.}
}
\label{fig:ant_map}
\end{center}
\end{figure}

Why default to observing failures when we don’t know what is going on? Because the success-failure boundary is full of information. Let me define ``failure" in the context of this discussion.
Imagine being an ant dropped somewhere on top of Fig.\ref{fig:ant_map}-(a). What is the fastest way to map your surroundings? Rather than walking every inch of the surface, find boundaries. When you are investigating a complex system that is working as expected, you are an ant dropped on Mount Success. To find the boundary, you have to push the system into failure mode. 
Staying inside the success space can inform you about the robustness of the system to perturbations (at best the system recovered from the perturbation, at worst your perturbation was irrelevant), but it is not explanatory. Neither is going from failure to failure. You can only investigate causes and effects if your intervention actually changes something: the failure boundary is not just more informative, it is a different kind of information altogether. 

Boundaries and failures are not exactly the same. If you are observing a function of the system that does not change when it crosses the failure boundary, you will not notice the transition. If you are observing the right function, you might see the system performance on that function become better or worse. Let us call ``failure points" abrupt transitions from “some performance” to 0: they are the most salient of transitions. Going back to Fig.\ref{fig:ant_map}, the ant might not notice the transition from a gentle slope to a flat terrain, but a cliff will be noticeable.
Ideally, you would want to map the entire failure boundary; in practice, you will focus on failure points.

\section{ Failure as a Fingerprint of Complex Systems}

\begin{figure}[!htb]
\begin{center}
\includegraphics[width=\linewidth]{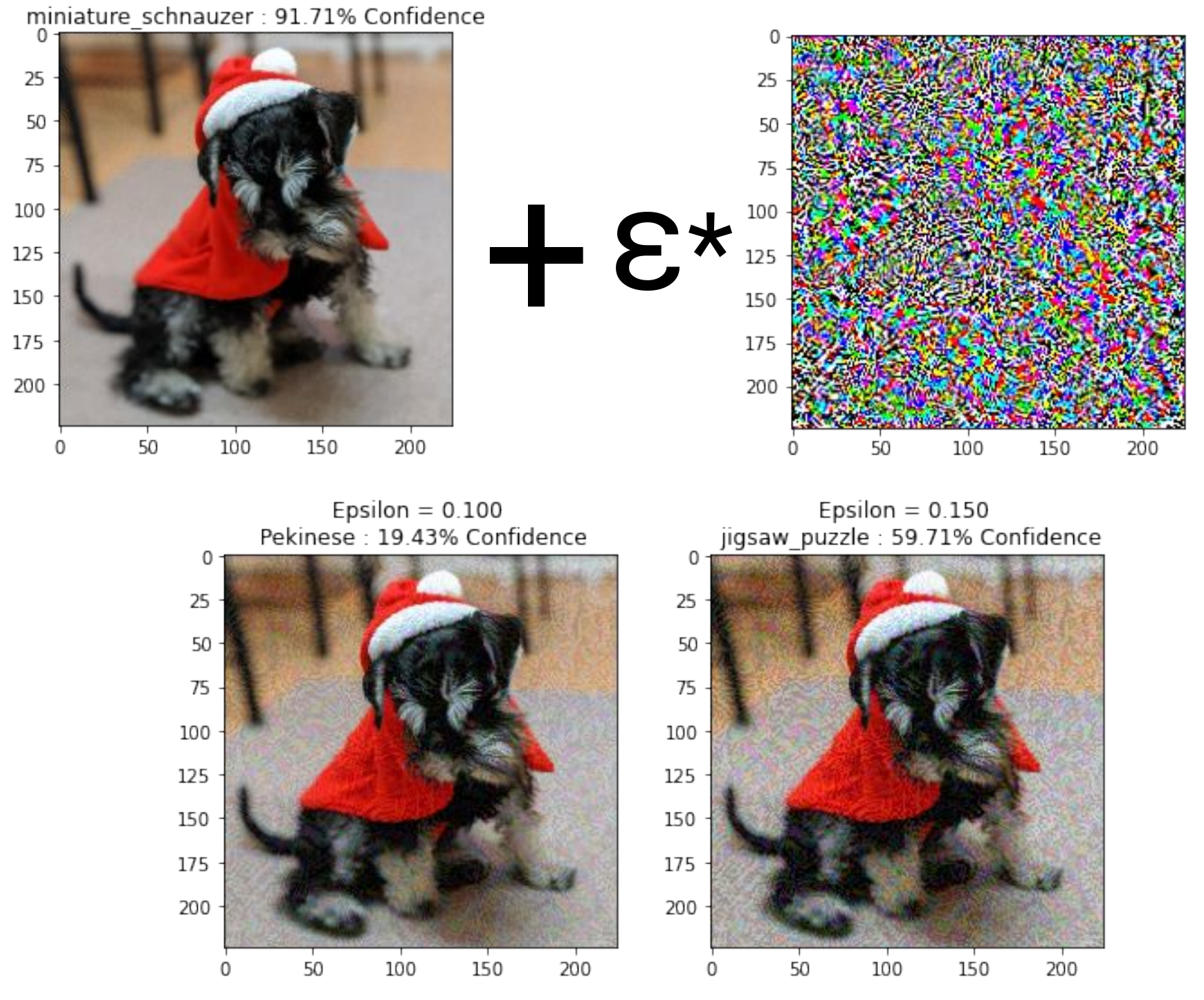}
\caption{\textbf{Artificial Neural Networks' mistakes look nothing like the mistakes of biological networks.} Non-random noise added to a picture of the author's dog (confusingly named ``Cookie") fools the network into misclassifying the image. Examples generated from \cite{adversarialTutorial} based on \cite{goodfellow2015explaining}}
\label{fig:cookie}
\end{center}
\end{figure}

The re-discovery of adversarial images (\cite{szegedy2014intriguing}, Fig.\ref{fig:cookie}) was met with dismay in the Deep Learning community, who had welcomed claims of ``super-human” performance at image recognition with comparatively little skepticism (\cite{stallkamp2012man, cirecsan2011committee}). Examining what the networks are doing ``better" than humans reveals that the comparison makes little sense: is a human wrong to label an image of 3 people in a golf cart as ``people", when the expected label is "kid"? Is a picture of a wall covered in pictures and papers best described by the official label "study"?\footnote{Examples found in ImageNet (\cite{deng2009imagenet}). Images could not be added to this manuscript for reasons of copyright and consent.} The artificial networks are better tuned to the dataset's idiosyncracies than to the structure of real world data (\cite{beyer2020we}).
The failures themselves are not what matters: the problem is the egregious difference between the artificial and biological system’s failures, despite initially overlapping successes that prompted some singularitarians\footnote{Believers in the ``singularity" claim that humans are about to fuse with a ``superintelligence"; Ray Kurzweil is the most famous singularity activist.} to claim that ``the brain achieves deep learning" (\cite{KurzweilDL}) and some AI researchers to wonder whether deep learning can be used to model brain processes (\cite{bengio2016biologically}). Some versions of this position might be true, but they are weakened by the fact that the failure boundaries of the two systems are so different. \cite{yamins2016using} propose deep hierarchical neural networks as  a path to ``produce quantitatively accurate computational models of sensory systems", yet also concede that adversarial images could be the sign of ``a fundamental architectural flaw in [deep hierarchical neural networks] as brain models". 

Two systems can succeed at a the same task yet succeed in completely different ways: planes are good for flying, but their flight is very different from bird flight. Conversely, two systems that fail differently at the same task are unlikely to be working in the same way, as is now recognized by AI researchers. For a given task, the failure boundary can be used as a fingerprint to identify systems that are genuinely similar in the way they perform (see Fig.~\ref{fig:ant_map}-(b)), even if the two systems are made of different ``stuff”. The fingerprint is substrate-agnostic.

Both engineering and natural sciences have embraced purposeful failure as a tool to deepen their understanding of complex systems. ALife and AI, as synthetic approaches, have the opportunity to use failure for even more than understanding existing biological systems: they can authenticate the models they build against the failure boundary of a biological target, to build (even better: evolve!) faithful models of those systems.

AI in particular strongly rejects failure, overwhelmingly focusing on what it perceives as glorifiable biological successes, 
especially successes rooted in anthropodenial. “[insert task here] is what distinguishes Man from the animal!” ...and the bulk of AI work to focus on chess, or natural language processing, or fuzzy logic. Beyond human exceptionalism, another powerful drive in AI is the reality of practical applications, where failure is very much undesirable\footnote{For example, the imitation of various types of bigotry by image or text processing networks is both undesirable and disappointing; it is also an unsurprising result, and therefore of low informative value.}. This results in an unbalanced availability of funding for short term applications, overshadowing theoretical work.

Recent work on visual illusions, including my own, provides examples going in the opposite, ``failure fingerprint" direction. A model of network trained to predict video frames was found to be sensitive to the same visual illusions that trick humans, while still treating regular images as non-illusions (\cite{watanabe2018illusory}). Illusions are a fascinating example of sensory failure; the fact that an artificial system can be fooled the same way, and even create its own illusions that fool humans~(\cite{eigen}), is a good example of shared failure point. Despite the black-box nature of the networks, this research tells us that interactions between our visual environment and our prediction abilities could be the root cause of visual illusions.

While AI is at the moment dominated by engineering successes, for better or for worse such is not the case for ALife. This on the positive side leaves a bigger place to fundamental research. For example, the search for minimal living cells in Wet ALife embraces biological failure as a research guide: a cell is said to be minimal if removal of any of its genetic information makes the cell unable to survive and reproduce (\cite{jean2019minimal, breuer2019essential}).
Failures are worth emulating for the sake of knowledge. Once this knowledge is acquired and we obtain causal understanding of a system through its reconstructed version, it is time to expand the reconstruction past its failures. After all, Artificial Life is about ``life as it could be": if we can emulate the failure fingerprint of some emblematic living systems, we have a path towards modifying this fingerprint, creating nontrivially strange forms of life.

\section{Failure for complexity}

\cite{oee} called Open Endedness “the last grand challenge you’ve never heard of”. Open Ended Evolution is one of ALife’s greatest ambitions. Contemporary Open Ended Evolution research focuses on obtaining the continual production of ``novelty" in artificial worlds: it is sometimes described as nontrivial, ever-increasing complexity. 
Desirable models are usually defined by lists of expected successes. Is it possible to replace them with lists of expected boundaries from success to failure? In that regard, Open Ended Evolution poses a challenge, as its definition is a list of boundless successes: boundless complexity, boundless novelty, boundless emergence...

Emergence, the process by which a group of units becomes more than the sum of its parts, is believed to be one of the mechanisms by which the nontrivial complexity of Open Endedness might be attained (\cite{banzhaf2016defining, stepney2016open} ).
When the ``parts” are living organisms such as swarms of birds, coral reefs, or social networks, repressing egoistical behavior (even to the point of giving up self reproduction), is a prerequisite for successful cooperation (\cite{szathmary2015toward, michod2001cooperation}).
For cooperation to persist, the group must have control mechanisms over egoistical individuals, imposing a cost on cheaters greater than the potential rewards. In other words, cooperation (and with it, emergence) requires the ability for a group to induce failure in any individual belonging to the group. This could be the ultimate example of top-down causation~(\cite{ellis2008nature, campbell1974downward}): a terminal control mechanism from an emergent scale over its individual components .

Take death, the ``failure of life”. In multicellular organisms, close cooperation between cells is occasionally broken by rogue mutant cells, that reproduce too much and refuse to die (\cite{zornig2001apoptosis}): cancer cells. Apoptosis (also called programmed cell death or cellular suicide), especially when provoked by external signals, is a mechanism for the organism to control cells that are abnormal or no longer needed (\cite{carson1993apoptosis}).
\cite{werfel2015} set out to find whether death could offer evolutionary benefits, and therefore evolve in a population of originally immortal organisms. Their conclusion was that death is not working against living systems, but offers a way to preserve ecosystems’ long term survival through environmental feedback, preventing over-exploitation of the resources of the environment. The ecosystem comprised of a spatial environment and its resident species self-regulates through death.

Over different spatiotemporal scales, failure offers a control mechanism that mediates what \cite{szathmary2015toward} called ``major transitions", of which they say that ``evolution at the lower level must be somehow constrained by the higher level". Failure boundaries, and in particular the boundary between life and death, can be engineered as control mechanisms. The more tightly knit a group becomes, the more evolutionary selection changes from being applied to individuals to being applied to the group, and the more internal control is favored. Cell destroys own proteins, organism kills own cells, group exiles organism, ecosystem destroys species. 
Open Ended Evolution's endless creativity requires prominent and accessible kill switches at every level, so that every unit of selection has a the power to timely get rid of its components. 

Beyond emulating the failure fingerprint of a simple life form, making it go through evolutionary transitions will require keeping its failure points accessible to others agents in the environment.



\section{Failure and Self-Directed Evolution}

``We need to invoke the capacity of organisms to pursue goals in order to explain the origin of adaptive novelties", said \cite{walsh2015organisms}, referring to the fact that a mutation needs to match the goals of an organism to be exploited by that organism. 
We can go further: failed behavior selects, before they appear, which mutations will be maintained in a population. It does more than just weeding out deleterious mutations after the fact.

Imagine a species of flies entirely guided by innate behavior in an environment with both food and poison. Through evolution, the species explores its behavioral space, recording what to eat and what to avoid as instinct in its DNA. Having sorted everything as either food or poison, it reaches the limits of successful behavior. Any deviation leads to failure.
Now sometime along its evolution, a mutation arises that makes one fly able to digest one type of red poison as harmless food. Unfortunately, if no fly ever tries to eat the red crumb, the new mutation might well disappear from the population without ever being used. A way around this missed opportunity is to keep a degree of adventurousness and failure in the behavior of individuals. If 1\% of all flies sometimes decide to eat the red poison (colorblindness? curiosity?), when the useful mutation appears it has a higher chance to spread in the population. Without the until-then failing behavior, that mutation was effectively neutral. To offer a benefit, it would have to appear right at the same time as a behavioral eat-red-poison mutation, or spread neutrally waiting for the right behavior.

So far it is simply a problem of evolutionary exploration versus exploitation. Now if from the beginning, before the mutation, a great percentage of flies eat red poison, behavior maintained maybe because the poison's effects are mild, or it is the same shade of red as a red food, the mutation will spread faster when it arises, and the flies will have effectively forced the direction of their own evolution. 

Are random mutations more likely to spread in a population, or are mutations more likely to be selected in advance by the behavior of the species? Take another example: antibiotic resistance in bacteria. Some bacteria produce antibiotics to protect their resources from competing species. Theoretically, these competitors can mutate and acquire the ability to survive antibiotics. Yet widespread resistance to antibiotics is a new phenomenon, tied to humans (over)use of antibiotics.
In a paper accompanied by a striking video, \cite{baym2016spatiotemporal} showed how highly resistant mutant bacteria lineages can die out while to less resistant lineages proliferate. The originally less resistant bacteria reached the entry of the antibiotic-enhanced space first, and got the opportunity to fail and try until they found the right mutation to cross over; the more resistant bacteria did not reach the more toxic environment fast enough to exploit their advantage. ``evolution is not always led by the most resistant mutants."~ (\cite{baym2016spatiotemporal} )
We are in the era of proliferation of resistant bacteria because resistance genes now confer a critical advantage in a human environment. How often do those ``reasons for mutations to stick around” originate in the species behavior itself rather than in extrinsic factors? Even with our use of antibiotics, bacteria would not exploit their resistance mutations if they did not ``try" to infect humans. 

In terms of learning, i.e. optimisation at the lifetime scale rather than the at evolutionary scale, failure is a good indicator of whether a system is still learning or whether if it has settled inside its success space and risks missing learning opportunities. A good example of using the failure boundary to tailor learning opportunities is the highly successful Generative Adversarial Networks (GANs, \cite{goodfellow2014generative}). GANs are composed of two networks that learn together by alternatively trying to beat each other. A generator network creates data, trying to pass this data as being part of an initial real dataset. A discriminator network tries to determine which data belongs to the initial dataset and which data was created by the generator.
When training goes well, the discriminator and generator improve together: as the discriminator becomes better at telling fake from genuine, the generator becomes better at creating convincing fakes. The threshold for failure is effectively negotiated between the networks. From this point of view, GANs are a form of curriculum learning, adapting the difficulty of the task to the ability of the student.

This suggests several possible avenues for AI and Soft ALife. For example, how about optimizing for performance variance in novelty search algorithms? Quality-diversity algorithms, used to find a variety of high performance ways to solve a task, require a measure of performance\footnote{Note that fitness in evolutionary algorithms is used as a cause, e.g. the measure used to decide the number of offspring of an individual, while in biology fitness is a consequence, e.g. the reproductive success of the individual.}) and a measure of diversity. It is notoriously difficult to define ``diversity". Performance variance could be used as a fitness function to sidestep the need for a definition. Performance differences reflect a diversity of ways to fail, i.e. the actual consequences of different behaviors rather than the objective difference between behaviors. Optimizing for performance variance would means optimizing for more diverse behavioral consequences, including not failing at all (as in classical performance optimization) and the opposite, low-performance behaviors, that might become useful if the agent or the environment change. Keeping as many values of these measures as possible in the population helps it stretch right to its failure boundary.

\section{Conclusion}
Focusing on imitating successes has the advantage of leaving room for many degrees and ways of succeeding. It also means that imitating a system's successes cannot always tell us how it works. Can focusing on failures lead to more faithful models without sacrificing general properties of biological systems to idiosyncratic details? It is unlikely that failure will become the new success, but I believe that it can at least open new directions of fundamental research, and even lead to improvement in mainstream applications.\\
~\\You open the mysterious box, take the object out, and start removing pieces until something breaks.
\footnotesize
\bibliographystyle{apa-good}
\bibliography{sample}

\end{document}